\documentclass{article}
\usepackage{iclr2025_conference,times}
\usepackage{hyperref}
\usepackage{url}
\usepackage{graphicx}
\usepackage{booktabs}
\usepackage{array}

\title{Implementation of AI in Precision Medicine}

\author{
\small
\begin{tabular}[t]{@{}l l l@{}}
\textbf{Göktuğ Bender} & \textbf{Samer Faraj} & \textbf{Anand Bhardwaj} \\
{\mdseries Desautels Faculty of Management} & {\mdseries Desautels Faculty of Management} & {\mdseries Desautels Faculty of Management} \\
{\mdseries McGill University} & {\mdseries McGill University} & {\mdseries McGill University} \\
{\mdseries Montreal, Canada} & {\mdseries Montreal, Canada} & {\mdseries Montreal, Canada} \\
\texttt{goktug.bender@mail.mcgill.ca} & \texttt{samer.faraj@mcgill.ca} & \texttt{anand.bhardwaj@mcgill.ca}
\end{tabular}
}

\begin{document}

\maketitle

\begin{abstract}
Artificial intelligence (AI) has become increasingly central to precision medicine by enabling the integration and interpretation of multimodal data, yet implementation in clinical settings remains limited. This paper provides a scoping review of literature from 2019–2024 on the implementation of AI in precision medicine, identifying key barriers and enablers across data quality, clinical reliability, workflow integration, and governance. Through an ecosystem-based framework, we highlight the interdependent relationships shaping real-world translation and propose future directions to support trustworthy and sustainable implementation.
\end{abstract}

\section{Introduction}

Traditional healthcare models have difficulty addressing the complexity of modern healthcare needs, particularly given the increasingly multimodal nature of health data spanning genetic, clinical, behavioral, environmental, and lifestyle information (Topol, 2023; Judge et al., 2024; Schouten et al., 2025). As precision medicine emerges as a promising solution for integrating multimodal data into healthcare, a new implementation strategy is necessary due to the complexity of existing healthcare structures and the extent of interdisciplinary collaboration that is now required (Tobias et al., 2023). Furthermore, the vast volume of data and the multi-faceted parameters involved present significant challenges (Ho et al., 2020). However, advancements in artificial intelligence (AI), which have the capacity to process complex information at large scales, have yielded promising results (Acosta et al., 2022; Soenksen et al., 2022). AI in precision medicine excels at monitoring and early detection, analyzing longitudinal patient data to identify subtle changes that signal disease onset or progression. It rapidly integrates and analyzes large, complex datasets that would be challenging for humans to interpret manually. 
Despite the success of proof-of-concept projects, most AI applications in precision medicine are characterized by limited real-world clinical adoption (Aristidou et al., 2022; Stenzinger et al., 2023). Several reviews have already documented the application of AI in specific fields such as radiology (e.g., Saxena et al., 2022), cardiology (e.g., Mohsen et al., 2023), and oncology (e.g., Bhinder et al., 2021). However, there is little research examining AI implementation in precision medicine with a focus on technical, ethical, and workflow aspects. In this scoping review, we aimed to synthesize and evaluate implementation related insights from the 2019–2024 literature on AI-enabled precision medicine, providing a comprehensive analysis of the current state of the art and practical pathways for implementation.

\section{Methods}

A comprehensive search was conducted in the Scopus database, yielding 698 articles. The search was limited to article titles, abstracts, and keywords, using the following terms: (“Precision Medicine” OR “Personalized Medicine”) AND (“AI” OR “Machine Learning”) OR “Artificial Intelligence”) AND (“Implementation” OR “Implementing” OR “Implemented” OR “Implementational.”). The search was refined by restricting results to articles, reviews, and conference papers published in English between 2019 and 2024 to capture the current AI wave. Additionally, the search was limited to subject areas related to healthcare. The initial search identified 698 articles. After title screening and full-text assessment, 108 articles were included in the analysis. Inclusion requirements were a focus on a) implementation, b) AI/ML and c) precision medicine. Papers were excluded if their focus was on the development of an AI system or the evaluation of its effectiveness. Only studies that focused on implementing AI were included.

\section{Results}

\subsection{Geographical and Specialty Distribution}

Most studies (88\%) reported data or discussed findings in a general or multi-country context, reflecting the predominance of general review articles rather than site-specific implementations. The largest group of papers (43\%) had authors from multiple countries followed by authors from the U.S. (19\%), with China (7\%) a distant third. 

The literature showed highest concentration in cross-cutting topics as well as oncology. Second rank attention was given to cardiology and pathology  (see Table~\ref{tab:specialty}).

\begin{table}[h]
\centering
\caption{Distribution of Literature Based on Specialty}
\label{tab:specialty}
\begin{tabular}{>{\raggedright}p{6cm}p{4cm}}
\toprule
\textbf{Specialty} & \textbf{\% of Literature} \\
\midrule
Cross-cutting & 18.02\% \\
Oncology & 18.02\% \\
Cardiology & 9.01\% \\
Pathology & 7.21\% \\
Critical Care, Pediatrics, Pharmacology & 4.5\% each \\
Psychiatry, Hematology, Neurology, Nephrology, Orthopedics, Radiology, Genomic Medicine, etc. & $<4\%$ each \\
Others (e.g., Surgery, Sleep Medicine, Rehabilitation, Dentistry) & $<1\%$ each \\
\bottomrule
\end{tabular}
\end{table}

\subsection{Emerging AI Technologies in Precision Medicine}

In oncology, studies reported that AI can facilitate individualized treatment through estimating future risk (e.g., of cancer recurrence; mortality) and can aid drug discovery using machine learning (ML) (Hamamoto et al., 2020; Chua et al., 2021). In cardiology, ML models were applied to interpret ECG data, predict arrhythmias, and assess cardiovascular risk from continuous physiological signals captured by wearable or implantable devices (Armoundas et al., 2024; Van Den Eynde et al., 2023). In radiology, AI is being used to enhance image interpretation, tumor segmentation, treatment response assessment, and integration of imaging data with genomic and clinical information to personalize diagnostics (Dankwa-Mullan \& Weeraratne, 2022). 

Some studies highlighted experimentations with AI-enabled infrastructures such as the use of digital twins, continuously updated virtual replicas of patients, integrated multimodal biological, clinical, and lifestyle data to simulate disease progression, forecast treatment response, and guide decision-making (Coorey et al., 2022; Vallée et al., 2024). Other studies reported on the development of: AI-driven clinical decision support systems that adapt to patient-specific contexts (Gearhart et al., 2020), multi-omics integration pipelines linking genomic, proteomic, and metabolomic profiles for individualized diagnostics (Voigt et al., 2021); and digital biomarkers derived from sensor data to enable early detection and continuous monitoring (Al-Anazi et al., 2024). Collectively, these approaches reflect a shift toward adaptive, data-driven infrastructures in precision medicine, where AI connects patient-specific data to dynamic therapeutic strategies.

\subsection{Data Quality, Validation, and Reproducibility}

Multiple studies highlighted that current datasets are often fragmented, single-institutional, and limited in demographic or clinical diversity, constraining model generalizability and reproducibility (Van Den Eynde et al., 2022; Langlais et al., 2023; Walsh et al., 2020; Walter et al., 2023). The lack of publicly accessible, multi-institutional data repositories was reported as a major barrier to developing robust and equitable AI models (Cascella et al., 2023, Jiang et al., 2020). Several papers noted that model optimization is secondary to data quality underscoring that heterogeneous, well-labeled datasets are crucial to reducing bias and improving clinical relevance (Chua et al., 2021; Vandenberk et al., 2023; Walter et al., 2023). Studies also emphasized the need for standardized protocols for model validation and replication. Rigorous internal and external validation, as well as transparent reporting of model parameters and performance metrics, were proposed to address issues of publication bias and reproducibility (Jacoba et al., 2021; Krittanawong et al., 2019). 

To help solve data quality issues, studies suggested developing systems to continuously monitor AI models’ performance, leveraging their ability to adapt and learn from new data (Walter et al., 2023); implementing standardized data curation pipelines with common ontologies and metadata schemas to improve interoperability and comparability (Chua et al., 2021); encouraging prospective data collection that includes underrepresented populations and real-world clinical settings to enhance model fairness and representativeness (Vandenberk et al., 2023); and mandating transparent reporting practices to support reproducibility (Cobanaj et al., 2024). 

\subsection{Clinical Reliability}

Multiple studies focused on the limited clinical reliability of existing AI tools, particularly when applied to patient-specific decision-making. Deep learning and ML models were frequently described as “black boxes,” where the decision logic remains opaque even to expert users, hindering clinician trust and widespread adoption (Angehrn et al., 2020; Vandenberk et al., 2023). Several authors recognized that while AI models can achieve high accuracy under experimental conditions, their predictions often lack traceability to clinical reasoning (Kuwaiti et al., 2023; Walter et al., 2023). Studies reporting on the deployment of models trained in one institutional context, and based on specific populations, were found to perform inconsistently when deployed at other health settings (Yang et al., 2023; Chua et al., 2021). 

Several authors suggested adopting Explainable AI (XAI) frameworks that provide interpretable reasoning paths to allow clinicians to understand why models produce specific predictions (Kuwaiti et al., 2023) and mandating external and temporal validation across heterogeneous patient cohorts and institutions to confirm reproducibility and detect model drift over time (Rogers et al., 2024; Walter et al., 2023).

\subsection{Integration with Clinical Workflow}

Most AI applications in precision medicine remain at the proof-of-concept or decision-support stage. Several studies identified integration challenges, particularly poor interoperability with existing health information systems and insufficient clinician involvement in model design (De Arizón et al., 2023; Jang et al., 2024). Moreover, studies describe physicians having difficulty trusting the output of AI models (Kuwaiti et al., 2023; Lanotte et al., 2023). Regarding pediatric healthcare, some parents expressed concern that AI-driven technologies could diminish their role in making decisions for their child, highlighting the need for clinical integration approaches that preserve shared decision-making (Sisk et al., 2020).

Across studies, authors emphasized the importance that AI should augment rather than replace human expertise. The proposed “human–AI loop” framework envisions clinicians using AI-generated insights to inform decision-making while maintaining final responsibility and epistemic authority (Malamateniou et al., 2021; Ryan et al., 2023; Dlugatch et al., 2024). Moreover, AI competency training should be incorporated into professional education to strengthen workforce readiness and support adoption (Malamateniou et al., 2021; Khalifa \& Albadawy, 2024). Authors also suggested establishing standardized, AI-ready data infrastructures that are interoperable across institutions (Horgan et al., 2019; Huang et al., 2023). However, while there were widespread calls for standardized AI-ready data formats, no consensus currently exists on what these standards should entail.

\subsection{Ethical, Regulatory, and Governance Considerations}

Ethical and governance concerns were among the most frequently cited barriers to clinical translation of AI in precision medicine. Several studies warned that algorithmic bias arising from non-representative or poorly curated training data could reinforce existing inequities in healthcare delivery and outcomes (Walsh et al., 2020; Walter et al., 2023). Models trained on homogenous datasets perform unevenly across demographic groups, emphasizing the need for inclusivity and transparency in dataset composition. Issues of data privacy, informed consent, and patient autonomy were also widely discussed. Many authors highlighted uncertainty surrounding secondary data use and the adequacy of existing consent models in contexts where patient data are continuously repurposed for AI development (Al-Anazi et al., 2024). A recurring theme across studies was the need for stronger governance and oversight mechanisms. 

With regulators such as the FDA and European Medicines Agency (EMA) developing AI-specific guidance for clinical use, several authors underscored the importance of algorithmic auditing and clearly defined liability frameworks to ensure accountability throughout the AI lifecycle (Al-Anazi et al., 2024; Walter et al., 2023). Other suggestions include developing dynamic consent models that allow patients to update data-sharing preferences over time (Tsopra et al., 2021) and creating unified regulatory guidelines to support the safe translation of AI-based medical tools from research to practice (Walter et al., 2023).

\section{Conclusion}

\begin{figure}[h]
\centering
\includegraphics[width=0.9\textwidth]{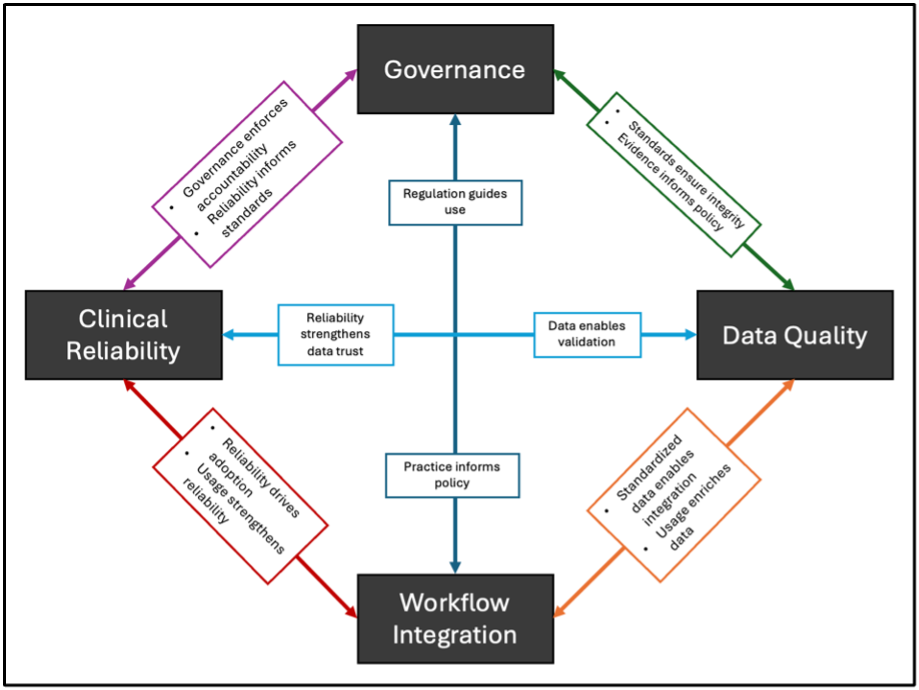} 
\caption{Ecosystem of Factors Influencing Implementation of AI in Precision Medicine.}
\label{fig:ecosystem}
\end{figure}

Across specialties, AI has demonstrated its potential to enable adaptive, personalized care, yet its translation into clinical implementation faces a multiplicity of limitations. This review highlights that the implementation of AI in precision medicine is not a linear progression from innovation to adoption, but a dynamic, interconnected ecosystem shaped by reciprocal relationships between governance, data quality, clinical reliability, and workflow integration. Based on this scoping review we offer a representation of the ecosystem of factors influencing the implementation of AI in precision medicine. (see Figure~\ref{fig:ecosystem}). The governance dimension in the model (e.g., issues of data integrity, transparency, and accountability) foregrounds the ethical, regulatory, and infrastructural foundation for trustworthy AI. The data quality dimension focuses on the importance of comprehensive, interoperable, and representative datasets. The clinical reliability dimension focuses on ensuring that AI models produce accurate, consistent, and clinically meaningful results under real-world conditions. The workflow integration dimension looks at physician confidence, usability within existing clinical systems and the extent to which AI tools complement rather than disrupt established clinical practices. 

The evidence generated from high-quality data and clinical practice continuously informs and refines governance standards. Data quality not only enable accurate model validation but also strengthen the reliability of AI outputs. In turn, clinically reliable models enhance physician confidence and promote seamless integration of AI tools into existing workflows. As these tools are deployed, real-world performance data, user feedback, and contextual constraints reveal gaps in data representativeness or model performance, feeding back into data governance and standard setting. These cyclical dependencies demonstrate that progress in one of the four domains depend on maturity in others. As such, data governance reforms must advance alongside rigorous model validation standards; ethical oversight should evolve in parallel with adaptive learning systems; and workflow integration must be guided by sustained interdisciplinary collaboration.

In conclusion, we propose that future research and policy initiatives recognize AI implementation as an ecosystem of interconnected enablers and barriers. Building trustworthy AI extends far beyond developing accurate algorithms; it necessitates coordinated investments in data quality and standardization, explainability frameworks, and clinician-AI collaboration that align computational reasoning with clinical judgment. As this review indicates, the challenges of AI enabled precision medicine are complex and not easily resolved. Nonetheless, there is optimism that through coordinated policy action, social engagement, and technological innovation, workable and sustainable solutions will emerge in the foreseeable future.

\subsubsection*{Acknowledgments}
This work was supported by a BMO Responsible AI Research Award awarded to Göktuğ Bender. Samer Faraj holds the Canada Research Chair in Technology, Innovation \& Organizing.

\subsubsection*{References}

{\small
\begin{itemize}
\setlength\itemsep{0.3em}
\setlength{\leftskip}{0pt}
\setlength{\itemindent}{0pt}
\setlength{\labelsep}{0.5em}
\setlength{\labelwidth}{1em}
\setlength{\listparindent}{0pt}

\item Acosta, J. N., Falcone, G. J., Rajpurkar, P., \& Topol, E. J. (2022). Multimodal biomedical AI. Nature Medicine, 28(9), 1773–1784. https://doi.org/10.1038/s41591-022-01981-2 

\item Al-Anazi, S., Al-Omari, A., Alanazi, S., Marar, A., Asad, M., Alawaji, F., \& Alwateid, S. (2024). Artificial intelligence in respiratory care: Current scenario and future perspective. Annals of Thoracic Medicine, 19(2), 117–130. https://doi.org/10.4103/atm.atm\_192\_23 

\item Angehrn, Z., Haldna, L., Zandvliet, A. S., Berglund, E. G., Zeeuw, J., Amzal, B., Cheung, S. Y. A., Polasek, T. M., Pfister, M., Kerbusch, T., \& Heckman, N. M. (2020). Artificial intelligence and machine learning applied at the point of care. Frontiers in Pharmacology, 11. https://doi.org/10.3389/fphar.2020.00759
\item Aristidou, A., Jena, R., \& Topol, E. J. (2022). Bridging the chasm between AI and clinical implementation. The Lancet, 399(10325), 620. https://doi.org/10.1016/s0140-6736(22)00235-5 
\item Armoundas, A. A., Narayan, S. M., Arnett, D. K., Spector-Bagdady, K., Bennett, D. A., Celi, L. A., Friedman, P. A., Gollob, M. H., Hall, J. L., Kwitek, A. E., Lett, E., Menon, B. K., Sheehan, K. A., \& Al-Zaiti, S. S. (2024). Use of artificial intelligence in improving outcomes in heart disease: a scientific statement from the American Heart Association. Circulation, 149(14). https://doi.org/10.1161/cir.0000000000001201 
\item Bhinder, B., Gilvary, C., Madhukar, N. S., \& Elemento, O. (2021). Artificial intelligence in cancer research and precision medicine. Cancer Discovery, 11(4), 900–915. https://doi.org/10.1158/2159-8290.cd-21-0090
\item Cascella, M., Montomoli, J., Bellini, V., Vittori, A., Biancuzzi, H., Mas, F. D., \& Bignami, E. G. (2023). Crossing the AI chasm in neurocritical care. Computers, 12(4), 83. https://doi.org/10.3390/computers12040083 
\item Chua, I. S., Gaziel‐Yablowitz, M., Korach, Z. T., Kehl, K. L., Levitan, N. A., Arriaga, Y. E., Jackson, G. P., Bates, D. W., \& Hassett, M. (2021). Artificial intelligence in oncology: Path to implementation. Cancer Medicine, 10(12), 4138–4149. https://doi.org/10.1002/cam4.3935
\item Cobanaj, M., Corti, C., Dee, E. C., McCullum, L., Boldrini, L., Schlam, I., Tolaney, S. M., Celi, L. A., Curigliano, G., \& Criscitiello, C. (2023). Advancing equitable and personalized cancer care: Novel applications and priorities of artificial intelligence for fairness and inclusivity in the patient care workflow. European Journal of Cancer, 198, 113504. https://doi.org/10.1016/j.ejca.2023.113504 
\item Coorey, G., Figtree, G. A., Fletcher, D. F., Snelson, V. J., Vernon, S. T., Winlaw, D., Grieve, S. M., McEwan, A., Yang, J. Y. H., Qian, P., O’Brien, K., Orchard, J., Kim, J., Patel, S., \& Redfern, J. (2022). The health digital twin to tackle cardiovascular disease—a review of an emerging interdisciplinary field. Npj Digital Medicine, 5(1). https://doi.org/10.1038/s41746-022-00640-7
\item Dankwa-Mullan, I., \& Weeraratne, D. (2022). Artificial intelligence and Machine learning Technologies in cancer Care: Addressing disparities, bias, and data diversity. Cancer Discovery, 12(6), 1423–1427. https://doi.org/10.1158/2159-8290.cd-22-0373  
\item De Arizón, L. F., Viera, E. R., Pilco, M., Perera, A., De Maeztu, G., Nicolau, A., Furlano, M., \& Torra, R. (2023). Artificial intelligence: a new field of knowledge for nephrologists? Clinical Kidney Journal, 16(12), 2314–2326. https://doi.org/10.1093/ckj/sfad182
\item Dlugatch, R., Georgieva, A., \& Kerasidou, A. (2023). Trustworthy artificial intelligence and ethical design: public perceptions of trustworthiness of an AI-based decision-support tool in the context of intrapartum care. BMC Medical Ethics, 24(1). https://doi.org/10.1186/s12910-023-00917-w 
\item Gearhart, A., Gaffar, S., \& Chang, A. C. (2020). A primer on artificial intelligence for the paediatric cardiologist. Cardiology in the Young, 30(7), 934–945. https://doi.org/10.1017/s1047951120001493 
\item Hamamoto, R., Suvarna, K., Yamada, M., Kobayashi, K., Shinkai, N., Miyake, M., Takahashi, M., Jinnai, S., Shimoyama, R., Sakai, A., Takasawa, K., Bolatkan, A., Shozu, K., Dozen, A., Machino, H., Takahashi, S., Asada, K., Komatsu, M., Sese, J., \& Kaneko, S. (2020). Application of Artificial Intelligence technology in Oncology: Towards the establishment of precision Medicine. Cancers, 12(12), 3532. https://doi.org/10.3390/cancers12123532  
\item Ho, D., Quake, S. R., McCabe, E. R., Chng, W. J., Chow, E. K., Ding, X., Gelb, B. D., Ginsburg, G. S., Hassenstab, J., Ho, C., Mobley, W. C., Nolan, G. P., Rosen, S. T., Tan, P., Yen, Y., \& Zarrinpar, A. (2020). Enabling technologies for personalized and precision medicine. Trends in Biotechnology, 38(5), 497–518. https://doi.org/10.1016/j.tibtech.2019.12.021
\item Horgan, D., Romao, M., Morré, S. A., \& Kalra, D. (2019). Artificial intelligence: power for civilisation – and for better healthcare. Public Health Genomics, 22(5–6), 145–161. https://doi.org/10.1159/000504785
\item Huang, J., Fan, X., \& Liu, W. (2023). Applications and Prospects of Artificial Intelligence-Assisted Endoscopic Ultrasound in digestive system diseases. Diagnostics, 13(17), 2815. https://doi.org/10.3390/diagnostics13172815 
\item Jacoba, C. M. P., Celi, L. A., \& Silva, P. S. (2021). Biomarkers for Progression in Diabetic Retinopathy: Expanding Personalized Medicine through Integration of AI with Electronic Health Records. Seminars in Ophthalmology, 36(4), 250–257. https://doi.org/10.1080/08820538.2021.1893351
\item Jang, S. J., Rosenstadt, J., Lee, E., \& Kunze, K. N. (2024). Artificial intelligence for Clinically Meaningful outcome Prediction in Orthopedic Research: Current applications and limitations. Current Reviews in Musculoskeletal Medicine, 17(6), 185–206. https://doi.org/10.1007/s12178-024-09893-z 
\item Jiang, Y., Yang, M., Wang, S., Li, X., \& Sun, Y. (2020). Emerging role of deep learning‐based artificial intelligence in tumor pathology. Cancer Communications, 40(4), 154–166. https://doi.org/10.1002/cac2.12012 
Judge, C. S., Krewer, F., O’Donnell, M. J., Kiely, L., Sexton, D., Taylor, G. W., Skorburg, J. A., \& Tripp, B. (2024). Multimodal artificial intelligence in medicine. Kidney360, 5(11), 1771–1779. https://doi.org/10.34067/kid.0000000000000556
\item Khalifa, M., \& Albadawy, M. (2024). Artificial intelligence for clinical prediction: Exploring key domains and essential functions. Computer Methods and Programs in Biomedicine Update, 5, 100148. https://doi.org/10.1016/j.cmpbup.2024.100148 
\item Krittanawong, C., Johnson, K. W., Rosenson, R. S., Wang, Z., Aydar, M., Baber, U., Min, J. K., Tang, W. H. W., Halperin, J. L., \& Narayan, S. M. (2019). Deep learning for cardiovascular medicine: a practical primer. European Heart Journal, 40(25), 2058–2073. https://doi.org/10.1093/eurheartj/ehz056 
\item Kuwaiti, A. A., Nazer, K., Al-Reedy, A., Al-Shehri, S., Al-Muhanna, A., Subbarayalu, A. V., Muhanna, D. A., \& Al-Muhanna, F. A. (2023). A review of the role of artificial intelligence in healthcare. Journal of Personalized Medicine, 13(6), 951. https://doi.org/10.3390/jpm13060951
\item Langlais, É. L., Thériault-Lauzier, P., Marquis-Gravel, G., Kulbay, M., So, D. Y., Tanguay, J., Ly, H. Q., Gallo, R., Lesage, F., \& Avram, R. (2022). Novel Artificial intelligence applications in Cardiology: current landscape, limitations, and the road to Real-World applications. Journal of Cardiovascular Translational Research, 16(3), 513–525. https://doi.org/10.1007/s12265-022-10260-x
\item Lanotte, F., O’Brien, M. K., \& Jayaraman, A. (2023). AI in Rehabilitation Medicine: Opportunities and Challenges. Annals of Rehabilitation Medicine, 47(6), 444–458. https://doi.org/10.5535/arm.23131 
\item Malamateniou, C., Knapp, K., Pergola, M., Woznitza, N., \& Hardy, M. (2021). Artificial intelligence in radiography: Where are we now and what does the future hold? Radiography, 27, S58–S62. https://doi.org/10.1016/j.radi.2021.07.015 
\item Mohsen, F., Al-Saadi, B., Abdi, N., Khan, S., \& Shah, Z. (2023). Artificial Intelligence-Based Methods for Precision Cardiovascular Medicine. Journal of Personalized Medicine, 13(8), 1268. https://doi.org/10.3390/jpm13081268 
\item Rogers, M. P., Janjua, H. M., Walczak, S., Baker, M., Read, M., Cios, K., Velanovich, V., Pietrobon, R., \& Kuo, P. C. (2023). Artificial intelligence in surgical Research: Accomplishments and future directions. The American Journal of Surgery, 230, 82–90. https://doi.org/10.1016/j.amjsurg.2023.10.045
\item Ryan, D. K., Maclean, R. H., Balston, A., Scourfield, A., Shah, A. D., \& Ross, J. (2023). Artificial intelligence and machine learning for clinical pharmacology. British Journal of Clinical Pharmacology, 90(3), 629–639. https://doi.org/10.1111/bcp.15930 
\item Saxena, S., Jena, B., Gupta, N., Das, S., Sarmah, D., Bhattacharya, P., Nath, T., Paul, S., Fouda, M. M., Kalra, M., Saba, L., Pareek, G., \& Suri, J. S. (2022). Role of artificial intelligence in radiogenomics for cancers in the era of Precision Medicine. Cancers, 14(12), 2860. https://doi.org/10.3390/cancers14122860 
\item Schouten, D., Nicoletti, G., Dille, B., Chia, C., Vendittelli, P., Schuurmans, M., Litjens, G., \& Khalili, N. (2025). Navigating the landscape of multimodal AI in medicine: A scoping review on technical challenges and clinical applications. Medical Image Analysis, 105, 103621. https://doi.org/10.1016/j.media.2025.103621
\item Sisk, B. A., Antes, A. L., Burrous, S., \& DuBois, J. M. (2020). Parental Attitudes toward Artificial Intelligence-Driven Precision Medicine Technologies in Pediatric Healthcare. Children, 7(9), 145. https://doi.org/10.3390/children7090145 
\item Soenksen, L. R., Ma, Y., Zeng, C., Boussioux, L., Carballo, K. V., Na, L., Wiberg, H. M., Li, M. L., Fuentes, I., \& Bertsimas, D. (2022). Integrated multimodal artificial intelligence framework for healthcare applications. Npj Digital Medicine, 5(1). https://doi.org/10.1038/s41746-022-00689-4 
\item Stenzinger, A., Moltzen, E. K., Winkler, E., Molnar‐Gabor, F., Malek, N., Costescu, A., Jensen, B. N., Nowak, F., Pinto, C., Ottersen, O. P., Schirmacher, P., Nordborg, J., Seufferlein, T., Fröhling, S., Edsjö, A., Garcia‐Foncillas, J., Normanno, N., Lundgren, B., Friedman, M., . . . Rosenquist, R. (2023). Implementation of precision medicine in healthcare—A European perspective. Journal of Internal Medicine, 294(4), 437–454. https://doi.org/10.1111/joim.13698
\item Tobias, D. K., Merino, J., Ahmad, A., Aiken, C., Benham, J. L., Bodhini, D., Clark, A. L., Colclough, K., Corcoy, R., Cromer, S. J., Duan, D., Felton, J. L., Francis, E. C., Gillard, P., Gingras, V., Gaillard, R., Haider, E., Hughes, A., Ikle, J. M., . . . Franks, P. W. (2023). Second international consensus report on gaps and opportunities for the clinical translation of precision diabetes medicine. Nature Medicine, 29(10), 2438–2457. https://doi.org/10.1038/s41591-023-02502-5 
\item Topol, E. J. (2023). As artificial intelligence goes multimodal, medical applications multiply. Science, 381(6663). https://doi.org/10.1126/science.adk6139
\item Tsopra, R., Fernandez, X., Luchinat, C., Alberghina, L., Lehrach, H., Vanoni, M., Dreher, F., Sezerman, O., Cuggia, M., De Tayrac, M., Miklasevics, E., Itu, L. M., Geanta, M., Ogilvie, L., Godey, F., Boldisor, C. N., Campillo-Gimenez, B., Cioroboiu, C., Ciusdel, C. F., . . . Burgun, A. (2021). A framework for validating AI in precision medicine: considerations from the European ITFoC consortium. BMC Medical Informatics and Decision Making, 21(1). https://doi.org/10.1186/s12911-021-01634-3 
\item Vallée, A. (2024). Challenges and directions for digital twin implementation in otorhinolaryngology. European Archives of Oto-Rhino-Laryngology, 281(11), 6155–6159. https://doi.org/10.1007/s00405-024-08662-5
\item Van Den Eynde, J., Kutty, S., Danford, D. A., \& Manlhiot, C. (2021). Artificial intelligence in pediatric cardiology: taking baby steps in the big world of data. Current Opinion in Cardiology, 37(1), 130–136. https://doi.org/10.1097/hco.0000000000000927 
\item Vandenberk, B., Chew, D. S., Prasana, D., Gupta, S., \& Exner, D. V. (2023). Successes and challenges of artificial intelligence in cardiology. Frontiers in Digital Health, 5. https://doi.org/10.3389/fdgth.2023.1201392
\item Voigt, I., Inojosa, H., Dillenseger, A., Haase, R., Akgün, K., \& Ziemssen, T. (2021). Digital twins for multiple sclerosis. Frontiers in Immunology, 12. https://doi.org/10.3389/fimmu.2021.669811 
\item Walsh, C. G., Chaudhry, B., Dua, P., Goodman, K. W., Kaplan, B., Kavuluru, R., Solomonides, A., \& Subbian, V. (2020). Stigma, biomarkers, and algorithmic bias: recommendations for precision behavioral health with artificial intelligence. JAMIA Open, 3(1), 9–15. https://doi.org/10.1093/jamiaopen/ooz054 
\item Walter, W., Pohlkamp, C., Meggendorfer, M., Nadarajah, N., Kern, W., Haferlach, C., \& Haferlach, T. (2022). Artificial intelligence in hematological diagnostics: Game changer or gadget? Blood Reviews, 58, 101019. https://doi.org/10.1016/j.blre.2022.101019 
\item Yang, J., Hao, S., Huang, J., Chen, T., Liu, R., Zhang, P., Feng, M., He, Y., Xiao, W., Hong, Y., \& Zhang, Z. (2023). The application of artificial intelligence in the management of sepsis. Medical Review, 3(5), 369–380. https://doi.org/10.1515/mr-2023-0039 

\end{itemize}
}

\end{document}